\documentclass[10pt]{article} 
\usepackage[preprint]{tmlr}

\usepackage{hyperref}
\usepackage{url}
\usepackage{lipsum}
\usepackage{amsmath,amsthm,amssymb}
\usepackage{graphicx}
\usepackage{hhline,url}
\usepackage{mathrsfs}
\usepackage{xcolor}

\usepackage{caption}
\usepackage{subcaption}

\usepackage{algorithm,algpseudocode}

\algnewcommand{\Initialize}[1]{%
  \State \textbf{Initialize:}
  \State \hspace*{\algorithmicindent}\parbox[t]{0.8\linewidth}{\raggedright #1}
}

\newtheorem{thm}{Theorem}[]
\newtheorem*{thm*}{Theorem}
\newtheorem{m-thm}[thm]{Meta-Theorem}
\newtheorem*{m-thm*}{Meta-Theorem}

\newtheorem{remark}{Remark}[]
\newenvironment{rem}{\begin{remark}\rm}{\end{remark}}
\newtheorem{prop}{Proposition}[]
\newtheorem*{prop*}{Proposition}
\newtheorem{Definition}{Definition}

\newtheorem{Corollary}[]{Corollary}

\newtheorem{Example}[]{Example}

\newtheorem{algor}[thm]{Method}

\newtheorem{Condition}[thm]{Condition}

\newtheorem{assp}{Assumption}[]

\newcommand{\Hil}{\mathcal{H}}

\newcommand{\R}{\mathbb{R}}
\newcommand{\dd}{\,\mathrm{d}}
\newcommand{\ve}{\varepsilon}

\renewcommand{\S}{\mathcal{S}}

\renewcommand{\phi}{\varphi}

\newcommand{\X}{\mathcal{X}}
\newcommand{\bm}[1]{{\mbox{\boldmath $#1$}}}

\newcommand{\EGP}[1]{\mathbb{E}_\mathcal{GP}\!\left[#1\right]}
\newcommand{\Epi}[1]{\mathbb{E}_{e\sim\pi}\!\left[#1\right]}

\newcommand{\ord}[1]{\mathcal{O}\!\left(#1\right)}

\newcommand{\A}{\mathcal{A}}
\newcommand{\ip}[1]{\left\langle #1 \right\rangle}

\newcommand{\vertiii}[1]{{\left\vert\kern-0.25ex\left\vert\kern-0.25ex\left\vert #1 
    \right\vert\kern-0.25ex\right\vert\kern-0.25ex\right\vert}}

\DeclareMathOperator{\wce}{wce}

\newcommand{\Z}{\mathcal{Z}}

\title{Policy Gradient with Kernel Quadrature}

\author{\name Satoshi Hayakawa \email hayakawa@maths.ox.ac.uk \\
      \addr Mathematical Institute,
      University of Oxford
      \AND
    \name Tetsuro Morimura \email morimura\_tetsuro@cyberagent.co.jp \\
      \addr CyberAgent, Inc.}

\def\month{MM}  
\def\year{YYYY} 
\def\openreview{\url{https://openreview.net/forum?id=XXXX}} 

\begin{document}

\maketitle

\begin{abstract}
    Reward evaluation of episodes becomes a bottleneck in a broad range of reinforcement learning tasks.
    Our aim in this paper is to select a small but representative subset of a large batch of episodes,
    only on which we actually compute rewards
    for more efficient policy gradient iterations.
    We build a Gaussian process modeling of discounted returns or rewards
    to derive a positive definite kernel on the space of episodes,
    run an ``episodic" kernel quadrature method
    to compress the information of sample episodes,
    and pass the reduced episodes to the policy network for gradient updates.
    We present the theoretical background of this procedure as well as its numerical illustrations in MuJoCo tasks.
\end{abstract}

\section{Introduction}\label{sec:intro}
Reinforcement learning (RL) aims to learn a policy model that maximizes the cumulative average of rewards \citep{Sutton2018rl}.
Policy gradient algorithms operate based on gradient ascent in the policy parameter space \citep{Gullapalli1990rl, Williams1992rl, Schulman2017rl} and have greatly benefited from the latest advancements in neural network models.
These algorithms have found widespread applications in domains such as robotics \citep{Peters2008rl}, large language models \citep{Ouyang2022rl}, medical diagnosis \citep{Xia2020rl}, and many others.

Despite the broad applications of RL, a significant challenge, often overlooked, is the extensive computational or monetary cost associated with reward evaluations in real-world RL scenarios.
Notably, in domains like material science and fluid dynamics, physical simulators are often used for evaluation of policy decisions \citep{Fan2020rl, Rajak2021rl}. 
These simulations tend to be computationally intensive. 
For example, reward calculation using flow simulation in \citet{Fan2020rl} requires about 1.1 hours for each episode.
Other computationally demanding tasks include Neural Architecture Search \citep{Zoph2017rl} and Ordering-Based Causal Discovery \citep{Wang2021rl}.
Furthermore, tasks like person re-identification \citep{Liu2019rl} and RL with Human Feedback (RLHF) \citep{Ouyang2022rl} demand human annotation for reward computation.
This human annotation process often becomes a bottleneck, emphasizing the need to minimize such instances. 
While recent approaches like RL with AI feedback (RLAIF) \citep{Lee2023rl} show promise, querying external AI typically incurs a cost, reinforcing the desire to minimize the number of reward evaluations.

Given this backdrop, our primary motivation is to alleviate the computational and monetary burdens of reward evaluations in the online RL setup.
We try to accelerate policy gradient methods by employing a novel approach that efficiently selects a representative subset of episodes for reward computations.

Let us start with introducing necessary notations in RL and policy gradient methods.

\subsection{MDP and policy}
We are given a state space $\S$, an action space $\A$,
a reward function $r:\S\times\A\to \R$, and transition probability
$p(\cdot| s, a)$ over $\S$ conditioned on each state-action pair
along with a distribution over initial states $\rho(s)$; the tuple of these defines a Markov decision process (MDP).
In the RL setting, it is typically the case that both the transition probabilities $p(\cdot| s, a)$ and the initial state distribution $\rho(s)$ are unknown to the agent, requiring the agent to learn an effective policy through interaction with the environment.

Stochastic policy
$\pi(\cdot|  s)$ is a probability density over $\A$ (with a canonical reference measure)
conditioned on each $s\in\S$. Given $p$ and $\pi$, we can generate a Markov chain
that starts from a state $s_0\in\S$ possibly drawn from a probability distribution associated with the MDP and continues as $a_t\sim \pi(\cdot|s_t)$
and $s_{t+1}\sim p(\cdot|s_t, a_t)$ for $t\ge0$.
We call such a Markov chain an {\it episode} $e=(s_t, a_t)_{t\ge 0}$;
it can be of finite length $T=T(e)$
due to some termination rule of MDP.
Since $p$ is fixed in our setting, we might abuse the notation as $e\sim\pi$
to represent the stochasticity of $e$.

Given an episode $e=(s_t, a_t)_{t\ge0}$,
the {\it discounted return} at time $t$
is defined as $R_t(e) := \sum_{u\ge t}\gamma^{u-t}
r(s_u, a_u)$, where $\gamma\in(0, 1)$ is a discount rate.
When the dependency on $e$ is apparent,
we might simply denote $r(e) = (r_t)_{t\ge0} = (r(s_t, a_t))_{t\ge0}$ and $R_t=\sum_{u\ge t}\gamma^{u-t}r_u$.
We finally define the $Q$-function $Q^\pi:\S\times\A\to\R$
and the value function $V^\pi:\S\to\R$ associated with the policy $\pi$ by
\[
    Q^\pi(s, a):=\mathbb{E}_{e\sim\pi}[R_0(e)|s_0=s, a_0=a],
    \qquad 
    V^\pi(s):=\mathbb{E}_{e\sim\pi}[R_0(e)|s_0=s]
    =\mathbb{E}_{a\sim\pi(\cdot|s)}[Q^\pi(s, a)].
\]

\subsection{Policy gradient}
With a parameterized stochastic policy $\pi_\theta$,
we aim to maximize $
    J(\pi_\theta):=\mathbb{E}_{e\sim\pi_\theta}[R_0(e)]
$
with respect to $\theta$.
Its gradient can be written as
\[
    \nabla_\theta J(\pi_\theta)
    = \mathbb{E}_{e\sim\pi}
    \left[\sum_{t\ge0}\gamma^t R_t(e) \nabla_\theta\log\pi_\theta(a_t|s_t)\right]
    = \mathbb{E}_{e\sim\pi}
    \left[\sum_{t\ge0}\gamma^t A^\pi(s_t, a_t)\nabla_\theta\log\pi_\theta(a_t|s_t)\right],
\]
where $A^\pi(s, a):=Q^\pi(s, a) - V^\pi(s)$ is the advantage function,
introduced here for variance reduction.

In practice,
we usually approximate the gradient by Monte Carlo integration by generating episodes
$e_1, \ldots, e_N \sim_\text{iid} \pi$:
\begin{align}
    \nabla_\theta J(\pi_\theta)
    &\approx \frac1N\sum_{i=1}^N \hat{G}(e_i), \label{eq:mc-grad}\\
    \hat{G}(e)&:=\sum_{t\ge0} \gamma^t \hat{A}_t(e)
    \nabla_\theta \log\pi_\theta(a_t|s_t),
    \label{eq:g-hat}
\end{align}
where $\hat{A}_t(e)$ is an approximation of $A^\pi(s_t, a_t)$,
{\it which requires evaluations of $r(s_t, a_t)$ for $t\ge0$}.
A typical choice for $\hat{A}$ is
$\hat{A}_t(e) = R_t(e) - V_\phi(s_t)$,
where $V_\phi$ is a parametric approximation of the value function, often referred to as the baseline and updated iteratively \citep{Williams1992rl}. We refer to this method as the vanilla policy gradient (vpg).

We shall denote the policy parameter by $\theta$ and all the other parameters
including baseline and kernel hyperparameters by $\phi$.
In order to highlight our motivation,
we assume that we can separate {\it running an episode $e$} and {\it evaluating the reward $r(e)$} (and so $\hat{A}_t(e)$); this policy gradient method is summarized in Algorithm~\ref{algo:vpg}.

\begin{algorithm}[H]
    \caption{Policy gradient}
    \label{algo:vpg}
    \begin{algorithmic}[1]
        \Require{A policy $\pi_\theta$, advantage estimator $\hat{A}$}
        \For{${\it iteration} = 1, 2, \ldots$}
        \State{Generate episodes $(e_i)_{i=1}^N\sim_{\mathrm{iid}}\pi_\theta$}
        \State{Compute $(r(e_i))_{i=1}^N$ and then $(\hat{A}(e_i))_{i=1}^N$}
        \State{$\theta \leftarrow \theta + \alpha N^{-1}\sum_{i=1}^N \hat{G}(e_i)$
        ($\alpha$: learning rate)}
        \State{Update $\hat{A}$ by using $(e_i, r(e_i))_{i=1}^N$}
        \EndFor
    \end{algorithmic}
\end{algorithm}

\subsection{Contribution}\label{sec:contribution}

To speed up the usual policy gradient methods, we propose combining policy gradient algorithms
and kernel quadrature for reducing episodes,
particularly aiming at expensive-reward situations
mentioned in Section~\ref{sec:intro}.

Kernel quadrature in this setting runs as follows:
given a positive definite kernel of episodes $K(e, e^\prime)$,
we approximate the $N$-point empirical measure of episodes
by a weighted $n$-point subset:
$\frac1N\sum_{i=1}^N \delta_{e_i} \approx \sum_{i\in I}w_i \delta_{e_i}$
with $I\subset\{1,\ldots, N\}$ and $\lvert I \rvert = n$.
This process is treated as a (black-box) function
$\mathrm{KQuad}(K, (e_i)_{i=1}^N)$
given by \citet{hayakawa21b}.
See Remark~\ref{rem:complexity}
for the computational complexity.

\begin{algorithm}[H]
    \caption{Vanilla PGKQ}\label{algo:kq-no-mean}
    \begin{algorithmic}[1]
        \Require{$n\ll N$, a policy $\pi_\theta$, advantage estimator $\hat{A}$,
        and episodic kernel $K$}
        \For{${\it iteration} = 1, 2, \ldots$}
        \State{Generate episodes $(e_i)_{i=1}^N\sim_{\mathrm{iid}}\pi_\theta$}
        \State{$I, (w_i)_{i\in I} \leftarrow \mathrm{KQuad}(K, (e_i)_{i=1}^N)$
        with $\lvert I\rvert=n$}
        \State{Compute $(r(e_i))_{i\in I}$ and then $(\hat{A}(e_i), \hat{G}(e_i))_{i\in I}$}
        \State{$\theta \leftarrow \theta + \alpha \sum_{i\in I} w_i \hat{G}(e_i)$}
        \State{Update $\hat{A}$ by using $(w_i, e_i, r(e_i))_{i\in I}$}
        \State{Update $K$ by using $(w_i, e_i, r(e_i))_{i\in I}$}
        \EndFor
    \end{algorithmic}
\end{algorithm}

The proposed algorithm is given in Algorithm~\ref{algo:kq-no-mean}
as {\it policy gradient with kernel quadrature} (PGKQ).
While the detailed explanation of kernel quadrature under an MDP
is described in Section~\ref{sec:kq-all},
our contributions are summarized as follows:
\begin{itemize}
    \item We develop a theory to make available the kernel quadrature over episodes,
        by introducing a Gaussian process modeling of returns or rewards,
        which can also be combined with policy gradient relatives
        including PPO \citep{Schulman2017rl}.
    \item We compare two versions of our PGKQ method with
    existing policy gradients with small and large batch sizes,
    and see the efficiency of PGKQ in catching up with the large-batch
    policy gradient only by small-batch reward observations.
\end{itemize}

\section{Related literature}
\subsection{Bayesian quadrature for policy gradient}
The most directly relevant to our study is the application of Bayesian quadrature \citep{oha91}
for estimating policy gradient \citep[BQPG;][]{gha07,gha16,ake21}.
They model the $Q$-function $Q^\pi$ by a Gaussian process (GP;
see Section~\ref{sec:kq-general} for a formal definition),
and estimate the policy gradient $\nabla_\theta J(\pi_\theta)$
as a posterior of a vector-valued GP
based upon (noisy) observations of $Q^\pi$.
Their main contribution from the viewpoint of gradient estimate
is placing a better weight than the uniform weight in \eqref{eq:mc-grad}.

Although we also use GP modeling and weighted gradient estimate,
our methods are different from these existing studies in the following two points.

\paragraph{Episodes reduction.}
        As already mentioned in Section~\ref{sec:contribution},
        our objective is to approximate a large batch of episodes
        by a smaller batch of weighted episodes
        in order to reduce the number of reward computations,
        while the variants of BQPG are on how to better estimate
        the policy gradient by a fixed batch of episodes.
        
\paragraph{Flexible kernel selection.}
    As is always the case with the use of Bayesian quadrature,
    we need to know the exact values of some integrals associated with the covariance kernel of the GP \citep[e.g., Eq.~(16) in][]{gha07} to execute BQPG.
    They need to use a specific class of kernel due to this limitation,
    while our method is valid for any choice of kernel because of
    the existence of a larger empirical measure that we want to approximate.

\subsection{Numerical integration in data science}
Estimating intractable integrals with a small number of integrand evaluations
classically ranges from
Monte Carlo \citep{met49}, cubature \citep{str71}
to QMC \citep{dic13},
while the recent literature also includes Bayesian/kernel quadrature \citep{oha91,che10,bac17},
recombination \citep{lit12,tch15},
coresets \citep{coreset17}, determinantal point processes \citep[DPPs;][]{bar20},
or dataset distillation \citep{d-dist18}.
It is impossible to explain each method in detail,
but they all agree on approximating a (probability) distribution,
which is typically a continuous distribution or large discrete data,
by a small (weighted) set (i.e., a discrete measure with small support).

Let us mention some existing applications of these methods towards
better/faster gradient estimates in data science,
stochastic gradient descents \citep[SGDs;][]{rob51} in particular. 
DPPs have already been applied to acquire better mini-batches for gradient estimates in SGDs
\citep{zha17,zha19,bar21},
while there also is an application of recombination with the same motivation \citep{cos20car}.
Small-GAN \citep{sin20} uses coresets for a better minibatch selection
when training generative adversarial networks \citep[GANs;][]{goo20}.
Kernel quadrature,
a method of seeking coreset based on kernel-based discrepancy,
has also been applied to iterative updates in data science;
minibatch selection in a warped Bayesian quadrature \citep{ada22} and Bayesian optimization
\citep{ada23,ada23b},
and dictionary compression in model-based RL \citep{cha23}.

Our contribution compared with these studies is on
the GP modeling specific to MDPs
and the resulting application of kernel quadrature
to suited for efficiently estimating the policy gradient.

\section{Kernel quadrature for reducing episodes}\label{sec:kq-all}

\subsection{Kernel quadrature in general}\label{sec:kq-general}
Kernel quadrature is a way of approximating a large or continuous distribution by a small discrete distribution.
For a positive definite kernel $K:\X\times\X\to\R$
and a probability distribution $\mu$, the aim of kernel quadrature is to find a good quadrature
rule $\mu_n = (w_i, x_i)_{i=1}^n$ --- a set of
weights $w_i\in\R$ and points $x_i\in\X$
that makes the following
the worst-case error small:
\[
    \wce(\mu_n; K, \mu):=\sup_{\lVert f\rVert_{\Hil}\le 1}
    \left\lvert
        \mu_n(f) - \mu(f)
    \right\rvert,
\]
where
$\Hil$ is the reproducing kernel Hilbert space (RKHS)
associated with the kernel $K$,
$\mu_n(f):=\sum_{i=1}^nw_if(x_i)$, and $\mu(f):=\mathbb{E}_{x\sim\mu}[f(x)]$.

There are many algorithms for this problem,
including but not limited to
a greedy algorithm called herding \citep{che10,hus12,bac12,tsu22},
weighted sampling methods \citep{bac17,bel19,bel20,bel21,epp23},
a subsampling methods called thinning \citep{dwi21,dwi22,she22};
see \citet[Table 1]{hayakawa21b} for a comparison of these methods.
We use the convex kernel quadrature \citep{hayakawa21b,hayakawa23nys}
for its empirical competence, but our method can incorporate
any kernel quadrature method feasible with a general kernel and a discrete space.

The choice of the kernel $K$ is essential for the applications of kernel quadrature
\citep{bri17},
and we propose
exploiting the covariance kernels of
GPs of the discounted returns $R_t(e)$ or
the reward function $r$ in Section~\ref{sec:gp-mdp},
in order to set an appropriate kernel over the space of episodes.

Formally, a real-valued GP on $\X$ is a distribution over the space of functions $\X\to\R$,
associated with a mean function $m:\X\to\R$ and a covariance kernel function $K:\X\times\X\to\R$, denoted as $\mathcal{GP}(m, K)$.
It satisfies that, for a random function $f\sim\mathcal{GP}(m, K)$
and any $x_1,\ldots,x_m\in\X$,
$(f(x_i))_{i=1}^m$ follows a normal distribution with
the mean vector $(m(x_i))_{i=1}^m$ and
the covariance matrix $(K(x_i, x_j))_{i,j=1}^m$.
We write the expectation with respect to the distribution of the Gaussian process
we have as $\mathbb{E}_\mathcal{GP}$ (e.g., $\EGP{f} = m$),
and call $f\sim \mathcal{GP}(m, K)$ {\it centered} if $m=0$ as a function.

Before going into details of the modeling in the case of MDP,
we give an important relation between kernel quadrature and GP.
The following is an immediate
generalization of \citet[Proposition~1]{hus12}:
\begin{prop}\label{prop:iso}
    Let $f\sim\mathcal{GP}(0, K)$ be a centered GP on $\X$
    and $\mu$ be a probability distribution satisfying
    $\int_\X K(x, x)\dd\mu(x) < \infty$.
    Then, for a kernel quadrature rule $\mu_n$ on the same space,
    we have
    \[
        \mathbb{E}_\mathcal{GP}[\left(\mu_n(f) - \mu(f)\right)^2]
        = \wce(\mu_n; K, \mu)^2.
    \]
\end{prop}
Note that the GP being centered is essential
since otherwise an additional term of
$(\mu_m(m) - \mu(m))^2$
with $m=\EGP{f}$ appears.
We give its proof in Section~\ref{sec:proof-iso} for completeness.
In the context of our methods,
$\mu$ is typically given by a large batch of episodes
and $\mu_n$ is a quadrature rule with a smaller support,
with which we estimate the policy gradient.
For more details on the role of this proposition,
see the following sections,
e.g., the description after Proposition~\ref{prop:ret-gp}.

\subsection{Gaussian process modeling of MDP}\label{sec:gp-mdp}
Recall the notations for MDPs introduced in Section~\ref{sec:intro}.
In particular, we are given a parametric policy $\pi=\pi_\theta$,
$\hat{A}_t(e)$ is an approximation of the advantage $A^\pi(s_t, a_t)$,
and $\hat{G}_t(e)$ is an episodic gradient defined by \eqref{eq:g-hat}.
Also, recall that our objective is to find a good subset of episodes {\it before}
evaluating rewards.
Our use of GP is based on the following heuristic/informal assumption:

\begin{assp}[informal]\label{assp:heu}
    For a set of episodes $(e_i)_{i=1}^N$ and weights $(w_i)_{i\in I}$
    with $I\subset\{1,\ldots,N\}$,
    if the {\it uncertainty} of
    $\frac1N\sum_{i=1}^N \sum_{t\ge0}\gamma^t\hat{A}_t(e_i)
        - \sum_{i\in I} w_i \sum_{t\ge0}\gamma^t\hat{A}_t(e_i)$
    is small, then the {\it uncertainty} of
    $\frac1N\sum_{i=1}^N \hat{G}(e_i) - \sum_{i\in I}w_i\hat{G}(e_i)$
    is also small.
\end{assp}

The {\it uncertainty} mentioned above is not a mathematical term,
but can artificially be introduced by using a GP.
By looking at \eqref{eq:g-hat},
we are assuming above that ignoring the vector-valued term $\nabla_\theta \log \pi_\theta(a_t|s_t)$
does not so much harm the quality of the reduced episodes.
This heuristic not only simplifies the derivation of kernels into the MDP context,
but also provides a unified treatment to similar policy-gradient methods such as
TRPO and PPO \citep{sch15,Schulman2017rl} where the loss functions are given by multiplying
the advantage function and other score-related terms.

Provided the above heuristic, what we need to do is
derive a GP for the following functional of an episode:
\begin{equation}
    \hat{F}(e) := \sum_{t\ge0}\gamma^t\hat{A}_t(e),
    \label{eq:f-hat}
\end{equation}
as we can then use kernel quadrature via Proposition~\ref{prop:iso}.
We want to directly model $\hat{F}$,
but will start by modeling smaller components such as $R_t$ and $r$
for more data and flexibility.

In the following, we introduce two ways of modeling $\hat{A}$
with an episodic GP.
These are based on a simpler {\it base GP} for either the return $R_t$ or a reward $r$.
Although we primarily consider the estimator $\hat{A}_t(e) = R_t(e) - V_\phi(s_t)$,
our argument can be generalized to more complicated estimators such as
$R_t(e) - V_\phi(s_t) - \gamma^{t_0-t}(R_{t_0}(e) - V_\phi(s_{t_0}))$
(for a certain $t_0$)
introduced by \citet{mni16}.

For simplicity of notation,
we shall write
\[
    z = (s, a), \ z_t = (s_t, a_t) \in \Z:=\S\times\A
\]
and $e = (s_t, a_t)_{t\ge0} = (z_t)_{t\ge0}$
for state-action pairs
in the following.
We present two ways of GP modeling in Sections~\ref{sec:ret-gp} \&
\ref{sec:rew-gp},
and we explain how to update the GPs (mean function, covariance kernel) over iterations
in Section~\ref{sec:update-gp}.

\subsubsection{Option 1: modeling $R_t$ with GP}\label{sec:ret-gp}
In this model, our base GP on $\Z$ is given by
\begin{equation}
    R_t|_{z_t = z}
    \sim \mathcal{GP}(V_\phi, k_\psi),
    \label{eq:ret-gp}
\end{equation}
where $V_\phi(z) := V_\phi(s)$ is the baseline function in the policy gradient algorithm and $k_\psi$ is a positive definite kernel
on the domain $\Z=\S\times\A$
with hyperparameter $\psi$.
It formally means,
for episodes $e=(z_t)_{t\ge0}$,
$e^\prime=(z_t^\prime)_{t\ge0}$
and time $t, u$,
we have ($\mathbb{C}\mathrm{ov}$ denotes covariance)
\begin{align}
    \EGP{R_t(e)} &= V_\phi(s_t), \label{eq:RGP-m}
    \\
    \mathbb{C}\mathrm{ov}_\mathcal{GP}[R_t(e)R_u(e^\prime)]
    &= k_\psi(z_t, z_u^\prime). \label{eq:RGP-v}
\end{align}

Our intuition behind this modeling is two-fold.
First, we
$Q^\pi\sim\mathcal{GP}(V_\phi, k_{0,\psi})$
to quantify the uncertainty of $Q^\pi$.
Second, we consider $R_t$
as an estimator of $Q^\pi(s_t, a_t)$,
whose variance is modeled by another independent GP,
i.e., $R_t|_{z_t=z}\sim\mathcal{GP}(Q^\pi(z), k_{1,\psi})$; this randomness is of a different kind from the above uncertainty,
since $R_t$ is still a random variable when fixing
a stochastic policy $\pi$ (or a stochastic environment),
while $Q^\pi$ is a deterministic function.
The modeling \eqref{eq:ret-gp} is obtained by combining
these GPs ($k_\psi = k_{0,\psi} + k_{1,\psi}$).

Given the modeling for $R_t$, the advantage estimator
$\hat{A}_t(e) = R_t(e) - V_\phi(s_t)$ follows
a centered GP of $z_t$ (and then the episode $e$) with a covariance kernel $k_\psi$
and we have the following kernel for $\hat{F}$.
Similar computations also apply to
other modeling such as $R_t(e) - V_\phi(s_t) - \gamma^{t_0-t}(R_{t_0}(e) - V_\phi(s_{t_0}))$.

\begin{prop}\label{prop:ret-gp}
    Suppose $(R_t)_{t\ge0}$ (as functions of an episode)
    follow a GP determined by \eqref{eq:RGP-m} and \eqref{eq:RGP-v}
    with $k_\psi$ being bounded.
    Then, the functional\footnote{Strictly speaking,
    we only justify that the sum in $\hat{F}$ converges in the $L^2(\mathcal{GP})$ space, when with infinite horizon; see Section~\ref{sec:proof-ret}.
    The same applies to Proposition~\ref{prop:rew-gp}.} $\hat{F}(e)=\sum_{t\ge0}\gamma^t(\hat{A}_t(e))$
    with $\hat{A}_t(e)=R_t(e)-V_\phi(s_t)$ follows a centered GP
    with a covariance kernel $K$ given by
    \begin{equation}
        K(e, e^\prime)
        = \sum_{t,u\ge0}\gamma^{t+u}k_\psi(z_t, z_u^\prime),
        \label{eq:ker-F-ret-gp}
    \end{equation}
    where $e=(z_t)_{t\ge0}$, $e^\prime = (z_u^\prime)_{u\ge0}$ are episodes.
\end{prop}
Given $\hat{F}\sim\mathcal{GP}(0, K)$ with \eqref{eq:ker-F-ret-gp},
we can apply Proposition~\ref{prop:iso} to see how kernel quadrature works.
Indeed, by letting $f= \hat{F}$ and $\mu=\frac1N\sum_{i=1}^N\delta_{e_i}$ in Proposition~\ref{prop:iso}, for a quadrature rule $\mu_n = (w_i, e_i)_{i\in I}$,
we have
\[
    \EGP{\Biggl(\frac1N\sum_{i=1}^N\hat{F}(e_i) - \sum_{i\in I} w_i\hat{F}(e_i)\Biggr)^2}
    = \wce(\mu_n; K, \mu)^2.
\]
Thus, provided Assumption~\ref{assp:heu},
a good kernel quadrature (i.e., that of a small worst-case error)
gives us a good gradient estimate $\sum_{i\in I}w_i\hat{G}(e_i)$
while the amount of actual reward computations is kept small.

The actual algorithm looks like Algorithm~\ref{algo:kq-no-mean},
where we first update $k_\psi$ (see Section~\ref{sec:update-gp})
and then $K$ as \eqref{eq:ker-F-ret-gp}.

\begin{rem}\label{rem:mean-zero}
    We could choose to use a parametric mean function 
    $m_\psi(z)$ instead of using $V_\phi(s)$
    in \eqref{eq:ret-gp} for more detailed modeling of $R_t$.
    However, then $\hat{F}$ could not represent the sum of advantage estimators
    and would not necessarily follow a centered GP,
    so we could not use Proposition~\ref{prop:iso}.
    One way to circumvent this issue is using a hybrid gradient estimate
    described in the following section.
\end{rem}

\subsubsection{Option 2: modeling $r$ with GP}\label{sec:rew-gp}
The object modeled by the GP in the previous section is not static over the iterations of the policy gradient
in that $R_t$ (and $Q^\pi$) is dependent on the policy $\pi$.
So the update of $k_\psi$ might be inaccurate
for future policies.

Our second option is thus modeling an static object, the reward function $r$:
\begin{equation}
    r\sim \mathcal{GP}(m_\psi, k_\psi),
    \label{eq:rew-gp}
\end{equation}
where $\psi$ is a hyperparameter for the mean function $m_\psi$
and the positive definite kernel $k_\psi$ on $\Z = \S\times\A$.

However, under the modeling \eqref{eq:rew-gp},
the estimator $\hat{A}_t = R_t - V_\phi(s_t)$ does not necessarily follow a centered GP
because $\mathbb{E}_{\mathcal{GP}}[\hat{A}_t(e)] = \sum_{u\ge t}\gamma^{u-t}m_\psi(z_u) - V_\phi(s_t)$.
As already pointed out in Remark~\ref{rem:mean-zero},
we cannot simply use Proposition~\ref{prop:iso}
to run an episodic kernel quadrature in this case.
Instead,
we can observe the following decomposition:
\begin{align}
    \nabla_\theta J(\pi_\theta)
    &=\Epi{\sum_{t\ge0}\gamma^t(R_t -b(s_t)) g_\theta(z_t)}\label{eq:original-gradient}\\
    &= \underbrace{\Epi{\sum_{t\ge0}\gamma^t(R_t - \EGP{R_t})g_\theta(z_t)}}_{
        \text{I: centered GP term}} + \underbrace{\Epi{\sum_{t\ge0}\gamma^t(\EGP{R_t} - b(s_t))g_\theta(z_t)}}_{
        \text{II: bias term}},
    \label{eq:gp-decomposition}
\end{align}
where $b:\S\to\R$ is any (integrable) baseline function
and $g_\theta(z_t) := \nabla_\theta\log\pi_\theta(a_t|s_t)$.

This decomposition can nicely be understood by introducing a {\it fake} reward $m_\psi$ (instead of $r$)
and the corresponding fake return
$R^\psi_t(e):=\sum_{t\ge0}\gamma^{u-t}m_\psi(z_u)
=\EGP{R_t(e)}$.
Indeed, the term II is given just by replacing
$R_t$ with $R^\psi_t$ in the right-hand side of \eqref{eq:original-gradient},
and the remaining term I works as a gradient modification.
Let us consider the context of reducing episodes
from a large batch $(e_i)_{i=1}^N$ to a weighted small batch $\mu_n=(w_i, e_i)_{i\in I}$.
Since we can compute the fake rewards/returns without access to $r$,
we can estimate the bias term II by using all the episodes $(e_i)_{i=1}^N$.

For the term I, now that the integrands $R_t - \EGP{R_t}$ are centered,
we can apply Proposition~\ref{prop:iso}
with the following representation of the GP for
the modified functional
$\hat{F}_\mathcal{GP}:=\sum_{t\ge0}\gamma^t(R_t-\EGP{R_t})$.

\begin{prop}\label{prop:rew-gp}
    Let the reward $r$ follows a GP given by \eqref{eq:rew-gp}
    with $m_\psi$ and $k_\psi$ being bounded.
    Then, the functional
    $\hat{F}_\mathcal{GP}(e)=\sum_{t\ge0}\gamma^t(R_t(e)-\EGP{R_t(e)})$
    follows a centered GP with a covariance kernel $K$ given by
    \begin{equation}
        K(e, e^\prime)
        = \sum_{t,u\ge0} (1+t)(1+u)\gamma^{t+u}k_\psi(z_t, z_u^\prime),
        \label{eq:ker-F-rew-gp}
    \end{equation}
    where $e=(z_t)_{t\ge0}$, $e^\prime = (z_u^\prime)_{u\ge0}$ are episodes.
\end{prop}
Let us now define $\hat{F}_b:= \sum_{t\ge0}\gamma^t(R_t-b(s_t))$
as a generalization of $\hat{F}$ in the previous sections
and $\hat{F}^\psi_b:=\sum_{t\ge0}\gamma^t(R^\psi_t-b(s_t))\ (=\hat{F}_b - \hat{F}_\mathcal{GP})$
be its fake counterpart.
Similarly to the previous section,
given episodes $(e_i)_{i=1}^N$ and $\mu=\frac1N\sum_{i=1}^N\delta_{e_i}$,
we can approximate $\frac1N\sum_{i=1}^N \hat{F}_b(e_i)$
by
$\sum_{i\in I}w_i \hat{F}_\mathcal{GP}(e_i)
+ \frac1N\sum_{i=1}^N \hat{F}_b^\psi(e_i)$,
where $\mu_n = (w_i, e_i)_{i\in I}$ is a quadrature rule,
given by running a kernel quadrature algorithm only regarding the term~I.
Note that this approximation is computable only with the reward evaluations
over episodes $(e_i)_{i\in I}$.

By using Proposition~\ref{prop:iso},
their mean squared error in terms of GP,
which is actually just about the error of $
\frac1N\sum_{i=1}^N \hat{F}_\mathcal{GP}(e_i) - 
\sum_{i\in I}w_i \hat{F}_\mathcal{GP}(e_i)$,
can again be represented
as $\wce(\mu_n; K, \mu)^2$, where $K$ is given by \eqref{eq:ker-F-rew-gp}
regarding the term I this time.

However, the gradient estimate is not as simple as
$\sum_{i\in I}w_i\hat{G}(e_i)$ in the previous section
(or Assumption~\ref{assp:heu}),
due to the use of a non-centered GP.
Under the notation of \eqref{eq:gp-decomposition},
let us define
\begin{align*}
    \hat{G}_b(e)&:=\sum_{t\ge0}\gamma^t(R_t(e) -b(s_t)) g_\theta(z_t),\\
    \hat{G}_\mathcal{GP}(e)&:=\sum_{t\ge0}\gamma^t(R_t(e) - \EGP{R_t})g_\theta(z_t),\\
    \hat{G}_b^\psi(e)&:=\sum_{t\ge0}\gamma^t(R_t^\psi(e) -b(s_t)) g_\theta(z_t),
\end{align*}
and then $\hat{G}_b^\psi\ (= \hat{G}_b - \hat{G}_\mathcal{GP})$ is computable without access to the true rewards.
By using these notations,
we replace $\frac1N\sum_{i=1}^N \hat{G}_b(e_i)$
with a lighter gradient estimate
$\sum_{i\in I}w_i \hat{G}_\mathcal{GP}(e_i)
+ \frac1N\sum_{i=1}^N \hat{G}_b^\psi(e_i)$.
The overall algorithm
is given by Algorithm~\ref{algo:kq}.
The way we update GP parameters is described in Section~\ref{sec:update-gp}.

\begin{algorithm}[H]
    \caption{PGKQ with non-centered GP}\label{algo:kq}
    \begin{algorithmic}[1]
        \Require{$n\ll N$, a policy $\pi_\theta$, baseline $b$,
        GP-mean $m_\psi$ and covariance $k_\psi$ modeling
        $r\sim\mathcal{GP}(m_\psi,k_\psi)$}
        \For{${\it iteration} = 1, 2, \ldots$}
        \State{Generate episodes $(e_i)_{i=1}^N\sim_{\mathrm{iid}}\pi_\theta$}
        \State Compute $K$ with \eqref{eq:ker-F-rew-gp}
        \State{$I, (w_i)_{i\in I} \leftarrow \mathrm{KQuad}(K, (e_i)_{i=1}^N)$
        with $\lvert I\rvert=n$}
        \State{Compute $(r(e_i))_{i\in I}$ and then $(\hat{G}_\mathcal{GP}(e_i))_{i\in I}$}
        \State $\theta \leftarrow \theta + \alpha (\sum_{i\in I} w_i \hat{G}_\mathcal{GP}(e_i)+ \frac1N \sum_{i=1}^N \hat{G}_b^\psi(e_i))$
        \State{Update $b$ by using $(e_i)_{i=1}^N$ and $m_\psi$}
        \State{Update $m_\psi, k_\psi$ by using $(w_i, e_i, r(e_i))_{i\in I}$}
        \EndFor
    \end{algorithmic}
\end{algorithm}

\begin{rem}\label{rem:fake-mdp}
    We propose using $b=V_\phi$ taken from the base
    policy gradient algorithm.
    We could update $V_\phi$ by
    using the weighted episodes $(w_i, e_i)_{i\in I}$
    like the updates of $m_\psi$
    (see Section~\ref{sec:update-gp}),
    but we actually propose updating $V_\phi$ based on fake rewards
    $m_\psi$,
    which clearly separates
    the roles of the policy gradient (PG) and
    kernel quadrature (KQ),
    since $\hat{G}_b^\psi$ is the usual PG-estimate given the fake rewards.
    Indeed, the KQ side then receives episodes $(e_i)_{i=1}^N$ and outputs a smaller batch of weighted episodes $(w_i, e_i)_{i\in I}$, while the PG side runs a usual PG with all the episodes $(e_i)_{i=1}^N$ associated with the fake reward $m_\psi$, with a modification of its gradient estimate by adding $\sum_{i\in I} w_i \hat{G}_\mathcal{GP}(e_i)$.
    
    By following this formulation, we can not only use $\hat{G}_b$ (or $\hat{G}_b^\psi$) but also incorporate any gradient estimator
    (in the PG-side for the {\it fake-reward} MDP) in Algorithm~\ref{algo:kq}.
    See also Appendix~\ref{sec:impl-pgkq} for more implementation details,
    including how we combine PGKQ with PPO.
\end{rem}

\subsubsection{Updating GP networks}\label{sec:update-gp}
Let us describe how we update $k_\psi$ (and $m_\psi$)
during the iterations of PGKQ.

\paragraph{Updating $k_\psi$.}
Although we should ideally update the GP based on the Bayes rule in a purely Bayesian setting, our GPs have to treat either a non-static target function (like $R_t$) or too many data points since each timestep in the MDP is a data point for the GP.
Following the maximum likelihood estimation (MLE)
and the deep kernel learning \citep{wil16} framework,
we set the loss function for $k_\psi$ as
\begin{equation}
    L_\mathrm{ker}(k_\psi)
    = \bm{y}^\top k_\psi(\bm{z}, \bm{z})\bm{y}
    + \log\det k_\psi(\bm{z}, \bm{z})
    \label{eq:ker-loss}
\end{equation}
and conduct the gradient descent over the iterations,
where $\bm{z}$ is given by arranging data points $z_t = (s_t, a_t)$ from all the episodes
$e\in (e_i)_{i\in I}$, which are {\it after} the reduction of kernel quadrature,
and $\bm{y}$ is the corresponding observed values,
such as $R_t - V_\phi(s_t)$ in Option~1 and $r(z_t) - m_\psi(z_t)$ in Option~2.
In practice, we conduct the gradient descent by splitting $(\bm{z}, \bm{y})$ into minibatches to avoid computing the full $k_\psi(\bm{z}, \bm{z})$.

\paragraph{Updating $m_\psi$ for non-centered GPs.}
We can either pass the optimization of $m_\psi$ into
the deep kernel learning by using the same loss in \eqref{eq:ker-loss},
or just independently learn $m_\psi$ by least-squares as we do in practice.
For the least-squares learning of $m_\psi$,
we can make use of the quadrature measure $\mu_n$ to incorporate the episode weights
as
\begin{equation}
    L_\mathrm{mean} = \mathbb{E}_{e\sim\mu_n}\left[
    \sum_{t\ge0}(1+t)\gamma^t(r_t - m_\psi(z_t))^2\right],
    \label{eq:mean-loss}
\end{equation}
where the discount factor $(1+t)\gamma^t$ comes from
\eqref{eq:ker-F-rew-gp}.

\begin{remark}[Computational Complexity]\label{rem:complexity}\rm
    Let us consider the overall complexity of the kernel quadrature with kernel learning
    part in PGKQ.
    In both options, the computation of the single
    $K(e, e^\prime)$ requires $T^2$-times
    computation of $k_\psi$ ($T$ is the episode length),
    and so, by combining with the kernel quadrature algorithm in \citet{hayakawa21b}
    (where we use all the episodes for Nystr{\"o}m's method),
    it takes $\ord{N^2(\log n + T^2) + n^2N\log(N/n)}$ to
    get an $n$-episode quadrature from an $N$-episode sample.
    If we set the batch size to $M$ in the kernel learning,
    then the kernel update takes $\ord{M^3}$ in each iteration
    and has $\ord{nT/M}$ iterations.
    Thus, the overall complexity
    for a single iteration of the PGKQ algorithm (for kernel quadrature and kernel learning)
    is given by $\ord{N^2(\log n + T^2) + n^2N\log(N/n) + nTM^2}$.
\end{remark}

\begin{figure*}[t]
    \centering
    \includegraphics[width=\hsize]{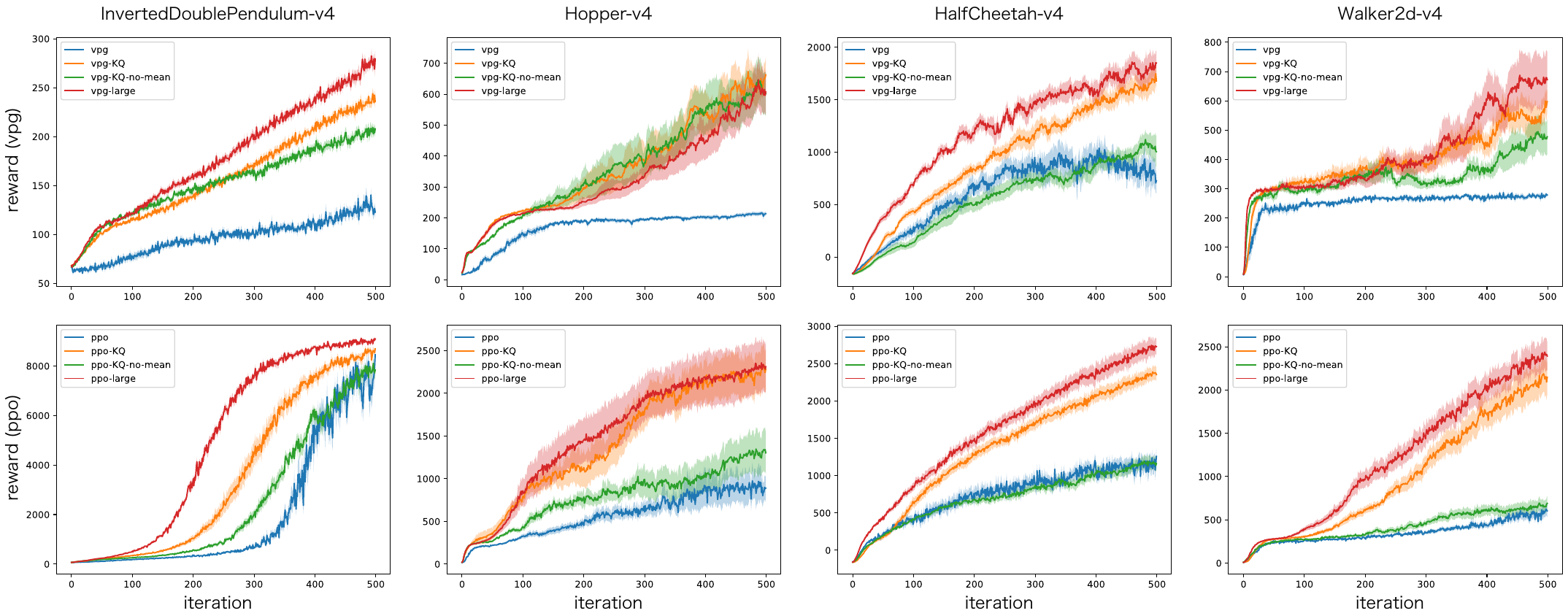}
    \caption{Learning curves in Mujoco tasks: reward vs iteration
    (comparing {\ttfamily \{vpg, ppo\}} (blue), {\ttfamily \{vpg, ppo\}-KQ} (orange),
    {\ttfamily \{vpg, ppo\}-KQ-no-mean} (green),
    and {\ttfamily \{vpg, ppo\}-large} (red))
    }
    \label{fig:mujoco-iter}
\end{figure*}

\section{Numerical experiments}\label{sec:experiment}
To demonstrate the effectiveness of our proposed methods, we conducted experiments on MuJoCo tasks since they are widely recognized as standard benchmarks in RL, even though the reward calculation for them is lightweight.

We evaluated PGKQ using two options, both designed to reduce a large batch of $N$ episodes to a smaller batch of $n$ episodes. We adopted two conventional PG methods, vpg and PPO, as the bases for PGKQ. We also compared these methods without kernel quadrature, using either small $n$ or large $N$ batch sizes. Throughout this section, we maintained $N=64$ and $n=8$. Specifically, we compared the following algorithms:
\begin{itemize}
    \item {\ttfamily \{vpg, ppo\}:}
    Corresponding algorithm with $n$ episodes per iteration.
    \item {\ttfamily \{vpg, ppo\}-KQ:}
    Corresponding algorithm with PGKQ by modeling $r$ (Option~2).
    \item {\ttfamily \{vpg, ppo\}-KQ-no-mean:}
    Corresponding algorithm with PGKQ by modeling $R_t$
    (Option~1).
    \item {\ttfamily \{vpg, ppo\}-large:}
    Corresponding algorithm with $N$ episodes per iteration.
\end{itemize}

In each experiment (either based on {\ttfamily vpg} or {\ttfamily ppo}), for each of the four algorithms, we ran a 500-iteration optimization (whose one iteration is based on
observing $n$ or $N$ episodes, depending on the algorithm)
10 times for statistical purposes.
In all the figures,
the empirical average of the total reward $\sum_{t\ge0}r(s_t, a_t)$
(which is actually a finite sum in all the experiments)
with its standard error (shaded region) are shown.
All the experiments were conducted by using PyTorch \citep{paszke2019pytorch} and Adam \citep{kingma2014adam}.

Our motivation for conducting these experiments is as follows:
{\it How well can we catch up with the learning curve of {\rm\ttfamily \{vpg, ppo\}-large}
by using PGKQ rather than the small-batch {\rm\ttfamily \{vpg, ppo\}}?}
So we compared the rewards against iterations
in this section,
but we also compare rewards against steps in Appendix~\ref{sec:rew-step}.

In all the experiments,
we used three-layer fully connected ReLU neural networks (NNs)
for each of $m_\psi$ and $k_\psi$,
where $k_\psi(z, z^\prime)$ was computed by
passing the NN-embeddings of state-action pairs $z$ and $z^\prime$
to the Gaussian kernel with additional scale and noise parameters.
See Appendix~\ref{sec:impl-pgkq} for implementation details.

\paragraph{MuJoCo tasks.}
We used MuJoCo \citep[v2.1.0,][]{mujoco} with the Gymnasium API \citep{towers_gymnasium_2023}\footnote{
All the experiments with MuJoCo were conducted with a Google Cloud Vertex AI notebook
with an NVIDIA T4
(16-core vCPU, 60 GB RAM).
}.
We compared all the aforementioned algorithms in the four following tasks:
{\ttfamily InvertedDoublePendulum-v4},
{\ttfamily Hopper-v4},
{\ttfamily HalfCheetah-v4},
{\ttfamily Walker2d-v4}.
In all the tasks, the maximum episode length is 1000,
where the tasks except {\ttfamily HalfCheetah-v4}
can terminate earlier when the state enters a predefined {\it unhealthy} region.

In these tasks, for the {\ttfamily vpg} (vanilla policy gradient) and
{\ttfamily ppo} \citep[proximal policy optimization with clipping,][]{Schulman2017rl},
we used the implementation of the machina\footnote{\url{https://github.com/DeepX-inc/machina}} library.
The learning rates of the policy, baseline, and GP-related networks were
all set to $3\times10^{-4}$.
This value was the default value of the machina library,
but also chosen as the most successful learning rate
when comparing several different learning rates with the 300-iteration, $n$-episode {\ttfamily ppo}
in the {\ttfamily Hopper-v4} task.
The discount rate was $\gamma=0.995$.

The results are shown in Figure~\ref{fig:mujoco-iter}.
In most tasks and base algorithms {\ttfamily \{vpg, ppo\}},
the order of total reward followed
{\ttfamily \{vpg, ppo\}-large $>$ \{vpg, ppo\}-KQ
$>$ \{vpg, ppo\}-KQ-no-mean $>$ \{vpg, ppo\}}.
In particular, in the experiments with {\ttfamily vpg},
PGKQ methods often get out of the plateau where
the {\ttfamily vpg} is stuck.


\section{Concluding remarks}
We have developed a method called PGKQ,
which combines existing policy gradient algorithm
with kernel quadrature over episodes,
aiming at better gradient estimates while keeping the number of actual reward computations small.
PGKQ is based on GP models of the return $R_t$ (Option~1)
or the reward function $r$ (Option~2),
whose covariance kernel accumulated over episodes gives
a kernel over the space of episodes,
which allows the use of kernel quadrature for selecting
a small but informative subset from a large batch of episodes.

We also confirmed the competitiveness of PGKQ, especially the Option~2,
by performing experiments in MuJoCo tasks.

As a future direction,
it should be beneficial to study a generalization of PGKQ based on
vector-valued GPs as in existing BQPG approaches \citep{gha07,gha16,ake21}
to avoid the heuristic Assumption~\ref{assp:heu}.

While we have been focusing on online RL, extending our approach to offline RL represents another interesting direction. In such a setup, determining which samples from a large batch are essential for reward computation, given the current learning policy, becomes critically important.

\bibliography{cite.bib}
\bibliographystyle{tmlr}

\appendix

\section{Proofs}
\subsection{Proof of Proposition~\ref{prop:iso}}\label{sec:proof-iso}
\begin{proof}
    Suppose $f\sim\mathcal{GP}(0, K)$.
    We first compute the following term:
    \begin{align*}
        \mathbb{E}_\mathcal{GP}[\mu(f)^2]
        = \mathbb{E}_\mathcal{GP}
        \biggl[\iint f(x)f(y)\dd\mu(x)\dd\mu(y)\biggr].
    \end{align*}
    By Cauchy-Schwarz, we have
    $\iint \lvert f(x)f(y)\rvert\dd\mu(x)\dd\mu(y) \le 
    (\int f(x)^2\dd\mu(x))^{1/2}(\int f(y)^2\dd\mu(y))^{1/2}
    = \int f(x)^2\dd\mu(x)$,
    so we have
    \[
        \EGP{\iint \lvert f(x)f(y)\rvert\dd\mu(x)\dd\mu(y)}
        \le \EGP{\int f(x)^2\dd\mu(x)}
        = \int\EGP{f(x)^2}\dd\mu(x)
        = \int K(x, x)\dd\mu(x) < \infty,
    \]
    where we have used Fubini's theorem (thanks to the nonnegativity of the integrand)
    in the first equality,
    and the assumption on the kernel in the last inequality.
    Thus, we can use Fubini's theorem for the original multiple integral,
    and we have, by using $\EGP{f(x)f(y)} = K(x, y)$,
    \[
        \EGP{\mu(f)^2}
        =\iint \EGP{f(x)f(y)}\dd\mu(x)\dd\mu(y)
        =\iint K(x, y)\dd\mu(x)\dd\mu(y).
    \]
    From the same use of Fubini's theorem
    by (partially) replacing $\mu$ with $\mu_n$,
    we can compute the other quadratic terms and obtain
    \begin{align*}
        \EGP{(\mu_n(f)-\mu(f))^2}
        = \iint K(x, y)\dd\mu(x)\dd\mu(y)- 2\sum_{i=1}^n w_i \int K(x_i, y)\dd\mu(y)
        + \sum_{i,j=1}^n w_iw_j K(x_i,x_j),
    \end{align*}
    which is a well-known formula for the $\wce(\mu_n; K, \mu)^2$
    \citep{gre06,sri10}.

    In general, we can also treat the non-centered case $f\sim\mathcal{GP}(m, K)$
    if $m$
    is integrable with respect to $\mu$.
    Indeed, we can apply the above computation to $f-m \sim\mathcal{GP}(0, K)$ to obtain
    \[
        \EGP{(\mu_n(f-m) - \mu(f-m))^2}
        = \wce(\mu_n; K, \mu)^2.
    \]
    From the linearity of the integral,
    we actually have
    \begin{align*}
        &\EGP{(\mu_n(f-m) - \mu(f-m))^2} \\
        &= \EGP{((\mu_n(f) - \mu(f)) - (\mu_n(m) - \mu(m)))^2} \\
        &= \EGP{(\mu_n(f)- \mu(f))^2}
        - 2(\mu_n(m)-\mu(m))\underbrace{\EGP{\mu_n(f) - \mu(f)}}_{=\mu_n(m) - \mu(m)} + (\mu_n(m)-\mu(m))^2\\
        &= \EGP{(\mu_n(f)- \mu(f))^2}
        - (\mu_n(m)-\mu(m))^2.
    \end{align*}
    Thus, in general, we have
    \[
        \EGP{(\mu_n(f) - \mu(f))^2}
        = \wce(\mu_n; K,\mu)^2 + (\mu_n-\mu(m))^2.
    \]
\end{proof}

\subsection{Proof of Proposition~\ref{prop:ret-gp}}\label{sec:proof-ret}
\begin{proof}
    From the modeling~\eqref{eq:RGP-m} and \eqref{eq:RGP-v},
    $\hat{A}_t$ formally follows a GP on the variables
    $(e, t)\in\mathcal{E}\times\mathcal{T}$ such that
    \[
        \EGP{\hat{A}_t(e)} = 0,
        \qquad
        \EGP{\hat{A}_t(e)\hat{A}_u(e^\prime)} = k_\psi(z_t, z^\prime_u),
    \]
    where $\mathcal{E}$ is the space of episodes
    and $\mathcal{T}:=\{0, 1,2,\ldots\}$ is the space of time indices.
    
    Let us define the space $L^2(\mathcal{GP})$
    as the space of $\R$-valued centered Gaussian variables
    given by taking the completion of the linear space
    $\mathop\mathrm{span}\{\hat{A}_t(e) \mid  (e,t)\in\mathcal{E}\times\mathcal{T}\}$
    with respect to the norm $\lVert X \rVert_{L^2(\mathcal{GP})}
    := \EGP{X^2}^{1/2}$.
    Then, since the kernel is bounded,
    we have
    $\sum_{t\ge0}\lVert \gamma^t\hat{A}_t(e)\rVert_{L^2(\mathcal{GP})}
    = \sum_{t\ge0}\gamma^tk(z_t, z_t)^{1/2} < \infty$,
    and so the infinite sum $\hat{F}(e)=\sum_{t\ge0}\gamma^t\hat{A}_t(e)$
    is well-defined in $L^2(\mathcal{GP})$.
    By letting $\hat{F}_T(e):=\sum_{t\ge0}^T\gamma^t\hat{A}_t(e)$,
    we have $\hat{F}_T\to\hat{F}$ in $L^2(\mathcal{GP})$.

    Therefore, we have that $\{\hat{F}(e)\mid e\in\mathcal{E}\}$
    is a family of centered (jointly) Gaussian variables with
    \begin{align}
        \EGP{\hat{F}(e)\hat{F}(e^\prime)}
        = 
        \lim_{T\to\infty}\lim_{U\to\infty} \EGP{\hat{F}_T(e)\hat{F}_U(e^\prime)}
        &=\lim_{T\to\infty}\lim_{U\to\infty}
        \sum_{t=0}^T\sum_{u=0}^U\gamma^{t+u}\EGP{\hat{A}_t(e)\hat{A}_{u}(e^\prime)}
        \nonumber\\
        &=\lim_{T\to\infty}\lim_{U\to\infty}
        \sum_{t=0}^T\sum_{u=0}^U\gamma^{t+u}k_\psi(z_t, z_u^\prime),
        \label{eq:prf-p2}
    \end{align}
    where $e=(z_t)_{t\ge0}$, $e^\prime = (z_u^\prime)_{u\ge0}$ are episodes.
    Since the kernel is bounded, the sum
    $\sum_{t,u\ge0}\gamma^{t+u}k_\psi(z_t, z_u^\prime)$ is absolutely convergent,
    and coincides with the right-hand side of \eqref{eq:prf-p2},
    which completes the proof.
\end{proof}

\subsection{Proof of Proposition~\ref{prop:rew-gp}}
\begin{proof}
    The flow of the proof is mostly the same as the previous one.
    The base GP is now $r\sim\mathcal{GP}(m_\psi, k_\psi)$.
    This time we define $L^2(\mathcal{GP})$
    as the space of $\R$-valued Gaussian variable given by the completion of
    the linear space $\mathop\mathrm{span}(\{1\}\cup\{r(z)\mid z\in\Z\})$
    with respect to the norm $\lVert X\rVert_{L^2(\mathcal{GP})} := \EGP{X^2}^{1/2}$.
    Since $m_\psi$ and $k_\psi$ are bounded,
    we first have that $R_t(e) = \sum_{u\ge t}\gamma^{u-t} r(z_u)$ is well-defined
    in $L^2(\mathcal{GP})$ as
    \[
        \sum_{u\ge t}\lVert \gamma^{u-t} r(z_u)\rVert_{L^2(\mathcal{GP})}
        = \sum_{u\ge t} \gamma^{u-t}\sqrt{m(z_u)^2 + k_\psi(z_u, z_u)}
        <\infty.
    \]
    Note that $L^2(\mathcal{GP})$ is a Hilbert space containing $1$
    with the inner product
    $\ip{X, X^\prime}_{L^2(\mathcal{GP})}=\EGP{XX^\prime}$.
    Thus, the expectation, which is the inner product with $1$,
    is continuous with respect to the norm,
    and so we have
    \[
        \EGP{R_t(e)} = \lim_{T\to\infty}\EGP{\sum_{u\ge t}\gamma^{u-t} r(z_u)}
        = \lim_{T\to\infty}\sum_{u=t}^T\gamma^{u-t}m_\psi(z_u) = \sum_{u\ge t}\gamma^{u-t}m_\psi(z_u),
    \]
    where the last infinite sum is
    absolutely convergent thanks to the boundedness of $m_\psi$.
    We can also prove that
    $R_t(e) - \EGP{R_t(e)} = \sum_{u\ge t} \gamma^{u-t} (r(z_u) - m_\psi(z_u))$
    by combining two convergent sequences.

    By letting $\hat{A}^{\mathcal{GP}}_t(e):=R_t(e) - \EGP{R_t(e)}$,
    we have a family of centered jointly Gaussian variables
    $\{\hat{A}^{\mathcal{GP}}_t(e)
    \mid (e,t) \in \mathcal{E}\times\mathcal{T}\}$ such that
    \[
        \lVert \hat{A}^\mathcal{GP}_t(e)\rVert_{L^2(\mathcal{GP})}
        \le \sum_{u\ge t}\lVert \gamma^{u-t} (r(z_u) - m_\psi(z_u)) \rVert_{L^2(\mathcal{GP})}\le
        \sum_{u\ge t} \gamma^{u-t} \sqrt{k_\psi(z_u, z_u)} < C
    \]
    for a constant $C>0$ independent of $t$ and $e$,
    where $\mathcal{E}$ and $\mathcal{T}$ are the spaces of episodes and time indices
    as defined in the previous proof.
    Thus, $\hat{F}_\mathcal{GP}(e) = \sum_{t\ge0} \gamma^t\hat{A}^\mathcal{GP}_t(e)$
    is well-defined in $L^2(\mathcal{GP})$ and
    $\{\hat{F}_\mathcal{GP}(e) \mid e\in\mathcal{E}\}$
    is a family of centered jointly Gaussian variables.
    Since the double sum
    $\sum_{t\ge0}\gamma^t\sum_{u\ge t}\lVert \gamma^{u-t} (r(z_u) - m_\psi(z_u)) \rVert_{L^2(\mathcal{GP})}$ is absolutely convergent,
    we can exchange the sum as
    \[
        \hat{F}_\mathcal{GP}(e)
        =\sum_{t\ge0}\gamma^t\sum_{u\ge t}\gamma^{u-t}(r(z_u)-m_\psi(z_u))
        =\sum_{u\ge0}\sum_{u=0}^t \gamma^u (r(z_u) - m_\psi(z_u))
        =\sum_{u\ge0}(1+u)\gamma^u (r(z_u) - m_\psi(z_u)).
    \]
    Therefore, the covariance kernel for $\hat{F}_\mathcal{GP}(e)$
    can formally be computed as
    \begin{align*}
        \EGP{\hat{F}_\mathcal{GP}(e)\hat{F}_\mathcal{GP}(e^\prime)}
        &= \sum_{t,u\ge0} (1+t)(1+u)\gamma^{t+u}
        \EGP{(r(z_t) - m_\psi(z_t))(r(z_u^\prime) - m_\psi(z_u^\prime))}\\
        &= \sum_{t,u\ge0} (1+t)(1+u)\gamma^{t+u}k_\psi(z_t, z_u^\prime)
    \end{align*}
    for episodes $e=(z_t)_{t\ge0}$, $e^\prime = (z_u^\prime)_{u\ge0}$,
    which is justified by the same logic as in the previous proof.
\end{proof}

\section{Experimental details}

\subsection{Implementation details}\label{sec:impl-pgkq}
\subsubsection{Combining policy gradient with kernel quadrature}
Recall that we have introduced in \eqref{eq:g-hat} the single-episode
gradient estimate
\begin{equation}
    \hat{G}(e)=\sum_{t\ge0} \gamma^t \hat{A}_t(e)
    \nabla_\theta \log\pi_\theta(a_t|s_t)
    \label{eq:app-ge}
\end{equation}
with an advantage estimator $\hat{A}_t$.
Although we have written that
we use $\frac1N\sum_{i=1}^N\hat{G}(e_i)$
as the Monte Carlo gradient estimate,
what we write in the actual code is
the computation of the (one-dimensional)
loss
\[
    (\text{Loss})=\frac1N \sum_{i=1}^N
    L_{\mathrm{vpg}}[\hat{A}_t](e_i),
    \quad
    \text{where} \quad 
    L_\mathrm{vpg}[\hat{A_t}](e):=\sum_{t\ge0}\gamma^t
    \hat{A}_t(e)\log\pi_\theta(a_t|s_t),
\]
and running an automatic differentiation
to get its gradient with respect to
the parameter $\theta$.
Here, $\hat{A}_t$ is treated as just a functional of an episode
in the definition of $L_\mathrm{vpg}[\hat{A}_t]$.
Let us explain how we actually compute the policy loss in the PGKQ given a kernel quadrature rule $\mu_n = (w_i, e_i)_{i\in I}$.
We start from VPG (vanilla policy gradient).
\begin{itemize}
    \item[(v1)]
        VPG with kernel quadrature with a centered GP
    ($\mathbb{E}_\mathcal{GP}[\hat{A}_t]=0$, Option~1) is the easiest case.
    Given $\mu_n$,
    we just replace $\frac1N \sum_{i=1}^N
        L_{\mathrm{vpg}}[\hat{A}_t](e_i)$
    with $\sum_{i\in I}w_i L_\mathrm{vpg}[\hat{A}_t](e_i)$
    in loss computation.
    \item[(v2)]
        When we combine VPG and a non-centered GP modeling of $r$ (Option~2),
        we just exploit
        the decomposition \eqref{eq:gp-decomposition} for loss computation
        (with a baseline function $b$):
        \begin{align*}
            (\text{Loss})&= \sum_{i\in I}w_i L_\mathrm{vI}(e_i) +
            \frac1N\sum_{i=1}^N L_\mathrm{vII}(e_i),
            \\
            &\text{where} \quad
            \begin{cases}
                L_\mathrm{vI}(e):=
                \sum_{t\ge0}\gamma^t
                (R_t(e) - \EGP{R_t(e)})\log\pi_\theta(a_t|s_t),\\
                L_\mathrm{vII}(e):=
                \sum_{t\ge0}\gamma^t
                (\EGP{R_t(e)} - b(s_t))\log\pi_\theta(a_t|s_t).
            \end{cases}
        \end{align*}
        As the baseline,
        we use the value estimator $V_\phi$
        trained with fake rewards as explained in Remark~\ref{rem:fake-mdp}.
        The observation $L_\mathrm{vI} = L_\mathrm{vpg}[R_t - \EGP{R_t}]$
        and $L_\mathrm{vII} = L_\mathrm{vpg}[\EGP{R_t}-b]$ allows a unified implementation.
\end{itemize}

We can also apply PGKQ to PPO \citep{Schulman2017rl}.
In the usual PPO,
we use the probability ratio (as a functional of an episode)
$q_t^\theta(e) :=\pi_\theta(a_t|s_t)/\pi_{\theta_\mathrm{old}}(a_t|s_t)$,
where $\theta_\mathrm{old}$ is the policy parameter
at which we assume the episodes $e_1,\ldots,e_N$ have been drawn.
Given an advantage estimator $\hat{A}_t$ as a functional
of an episode,
we compute the loss as follows:
\[
    (\text{Loss})=\frac1N\sum_{i=1}^N L_\mathrm{ppo}[\hat{A}_t](e_i),
    \quad \text{where} \quad
    L_\mathrm{ppo}[\hat{A}_t](e):= \sum_{t\ge0} \gamma^t \min\{
        q_t^\theta(e) \hat{A}_t(e),\ \mathrm{clip}(q_t^\theta(e), 1-\ve, 1+\ve)\hat{A}_t(e)\}.
\]
Here, $\mathrm{clip}(a,b,c):=\min\{\max\{a, b\}, c\}$
and $\ve$ ($=0.2$ in the implementation) is a clipping parameter.

\begin{itemize}
    \item[(p1)]
        PPO with the centered GP modeling (Option~1) is given by a straightforward replacement of $L_\mathrm{vpg}$ by
        $L_\mathrm{ppo}$
    \item[(p2)]
        When combining PPO with the Option~2,
        we can also imitate the decomposition of (v2):
        \[
            (\mathrm{Loss}) = \sum_{i\in I}w_iL_\mathrm{pI}(e_i) + \frac1N\sum_{i=1}^NL_\mathrm{pII}(e_i),
            \quad \text{where} \quad
            L_\mathrm{pI}:=L_\mathrm{ppo}[R_t-\EGP{R_t}],\ L_\mathrm{pII}:=L_\mathrm{ppo}[\EGP{R_t} - b].
        \]
        Here, $b$ is again the value estimator $V_\phi$ trained by fake rewards.
\end{itemize}
We can also consider the use of other advantage estimators than $R_t - b$.
Indeed,
$\hat{A}^\prime_t(e) = R_t(e) - V_\phi(s_t) - \gamma^{t_0-t}(R_{t_0}(e) - V_\phi(s_{t_0}))$ with $t_0$ being the final time
was adopted in the MuJoCo tasks (also following the original code of Machina).
The modification of (v1) and (p1) is straightforward as they formally do not depend on the specific
form of $\hat{A}_t$.
For (v2) and (p2), though there are other possibilities,
we just replaced $L_\mathrm{vII}$ and $L_\mathrm{pII}$ with
$L_\mathrm{vII}^\prime := L_\mathrm{vpg}[\mathbb{E}_\mathcal{GP}[\hat{A}_t^\prime]]$
and $L_\mathrm{pII}^\prime:= L_\mathrm{ppo}[\mathbb{E}_\mathcal{GP}[\hat{A}_t^\prime]]$, where
$\mathbb{E}_\mathcal{GP}[\hat{A}_t^\prime](e):=
\EGP{R_t(e)} - V_\phi(s_t) - \gamma^{t_0-t}(\EGP{R_{t_0}(e)} - V_\phi(s_{t_0}))$
is regarded as a functional of an episode.

\subsubsection{Network architecture}
For the policy and baseline networks, we used
the default implementations the library (Machina).
Let us explain the part with our original implementation,
$m_\psi$ and $k_\psi$ in the GP modeling.

\paragraph{Inputs.}
In all the MuJoCo tasks, we simply concatenated two vectors $s_t$ and $a_t$ to make $z_t$,
and used it as inputs to our networks.
Let us denote by $D$ the dimension of $z_t$ in the following.

\paragraph{Implementation of $k_\psi$.}
In both experiments,
we implemented $k_\psi$ as follows
\[
    k_\psi(z_t, z_u^\prime) = \exp\left(\lambda_\psi - \frac{\lVert f_\psi(z_t) - f_\psi(z_u^\prime)\rVert^2}{20}\right)
    + (10^{-5} + \exp(\sigma_\psi))\delta(z_t, z_u^\prime),
\]
where
$\lVert\cdot\rVert$ is the Euclidean norm,
$\delta(z_t, z_u^\prime)$ equals $1$ if $z_t=z_u^\prime$ and $0$ otherwise,
$\lambda_\psi, \sigma_\psi\in\R$ are respectively the scale and noise parameters,
and $f_\psi:\R^D\to\R^{10}$ is the embedding function explained in the following.

For $f_\psi$ in all the MuJoCo tasks,
we used a fully connected neural network with two hidden layers with $D$
(so $\R^D\to\R^D\to\R^D\to\R^{10}$)
and the ReLU activations except for the final layer.

\paragraph{Implementation of $m_\psi$.}
We used a fully connected neural network with two hidden layers with dimensions $200$ and $100$
(so $\R^D\to\R^{200}\to\R^{100}\to\R^{10}$)
and the ReLU activations except for the final layer.

\subsection{Reward vs step}\label{sec:rew-step}
\begin{figure}[H]
    \centering
    \includegraphics[width=\hsize]{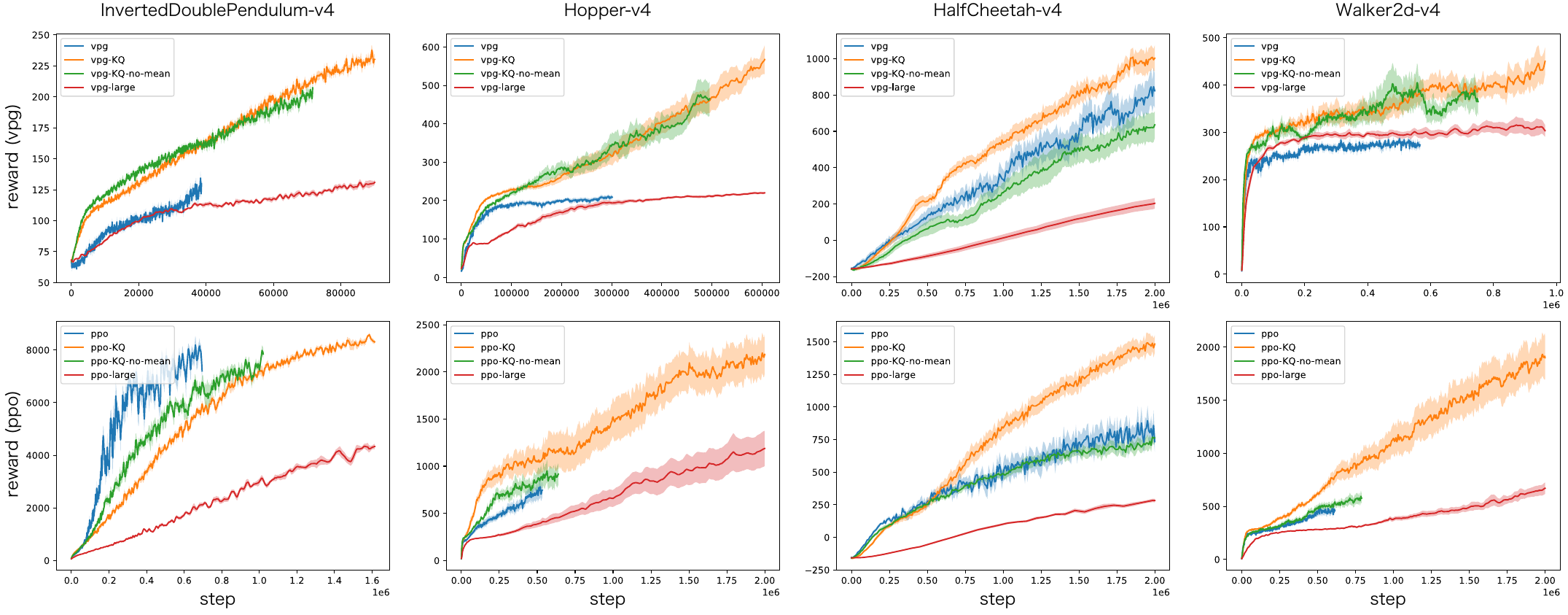}
    \caption{Learning curves in Mujoco Tasks: reward vs step}
    \label{fig:mujoco-step}
\end{figure}

Though we only compared the rewards against iterations in the main body,
since it is common to compare the rewards against observed steps \citep[e.g.,][]{Schulman2017rl},
we here present our experimental results in that regard.
Figure~\ref{fig:mujoco-step} is on each of the four MuJoCo tasks corresponding
to Figure~\ref{fig:mujoco-iter}.
Note that this is not necessarily a fair comparison to {\ttfamily\{vpg,ppo\}-large}
in terms of the learning rate per step.

\end{document}